\documentclass[letterpaper]{article} 
\usepackage{aaai2026}
\usepackage{times}  
\usepackage{helvet}  
\usepackage{courier}  
\usepackage[hyphens]{url}  
\usepackage{graphicx} 
\usepackage{natbib}  
\usepackage{caption} 
\usepackage{algorithm}
\usepackage{algorithmic}
\usepackage{amsmath}
\usepackage{booktabs}
\usepackage{newfloat}
\usepackage{listings}
\DeclareCaptionStyle{ruled}{labelfont=normalfont,labelsep=colon,strut=off} 
\floatstyle{ruled}
\newfloat{listing}{tb}{lst}{}
\floatname{listing}{Listing}
\title{BOOST: Bootstrapping Reasoning Programs for Program-Guided Fact-Checking}
\author{
    Qisheng Hu,
    Quanyu Long,
    Wenya Wang
}
\affiliations{
    Nanyang Technological University

}

\usepackage{bibentry}

\begin{document}

\maketitle

\begin{abstract}
Large language model (LLM) pipelines have improved automated fact-checking for complex claims, yet many approaches rely on few-shot in-context learning with demonstrations that require substantial human effort and domain expertise. Among these, program-guided reasoning—by decomposing claims into function calls and executing reasoning programs—has shown particular promise, but remains limited by the need for manually crafted demonstrations. Fundamentally, the underlying principles of effective reasoning program generation still remain underexplored. In this work, we introduce BOOST, a bootstrapping approach for automated few-shot reasoning program generation. BOOST iteratively refines explicit, data-driven guidelines as meta-rules for guiding demonstration creation, using a critique-refine loop that eliminates the need for human intervention. This enables a seamless transition from zero-shot to few-shot program-guided learning, enhancing interpretability and effectiveness. Experimental results show that BOOST outperforms prior few-shot baselines in both zero-shot and few-shot settings for complex claim verification.\footnote{Code: https://github.com/qishenghu/BOOST}
\end{abstract}

\section{Introduction}

\begin{figure*}[ht!]
    \centering
    \includegraphics[width=0.95\textwidth]{fig/pipeline_exp_placeholder.png}
    \caption{Overview of BOOST (left), which integrates explicit guidelines to guide reasoning program generation. BOOST employs a bootstrapping process to iteratively refine and generate guideline-driven demonstrations. An example partial response is illustrated (right), demonstrating the flexible composition of functions and the derivation of the \texttt{final\_prediction} variable, which serves as the system's final verification.}
    \label{fig:pipeline}
\end{figure*}

Automated claim verification is crucial for combating misinformation~\citep{guo2022survey}. Complex claim verification typically requires gathering multiple evidence pieces and multi-step reasoning~\citep{pan-etal-2023-fact,chen-etal-2024-complex}. In real-world scenarios, critical evidence is often scattered across documents, requiring evaluation of various claim aspects~\citep{chen-etal-2022-generating}. Additionally, multi-hop claims~\citep{jiang-etal-2020-hover,si-etal-2024-checkwhy} demand intermediary inferences before reaching a final verdict, further complicating the verification process.

To tackle this task, fact-checking frameworks increasingly adopt large language model (LLM)-based approaches that integrate decomposition, retrieval, and verification modules~\citep{chern2023factool,min-etal-2023-factscore,pan-etal-2023-fact} to jointly enhance performance, explainability, and data efficiency. Among these, \citet{pan-etal-2023-fact} pioneer program-guided reasoning, leveraging few-shot in-context learning (ICL). In this approach, the LLM is prompted with demonstrative reasoning programs—exemplars with claim-program pairs—to transform claims into reasoning programs with function calls. By delegating complex operations to reliable functions, program-guided reasoning enables LLMs to focus on higher-level symbolic reasoning while incorporating a formal execution layer that improves overall performance.

While program-guided reasoning has shown considerable promise for fact-checking, the underlying principles for generating effective reasoning programs remain underexplored. This gap poses significant challenges for designing few-shot demonstrations in ICL, as existing approaches~\citep{pan-etal-2023-fact, wang-shu-2023-explainable} depend on manually crafted examples that demand extensive domain expertise and require redesign for domain shifts. Automating effective reasoning program generation is thus challenging, requiring a deep understanding and formalizing the underlying logical structures that govern effective reasoning programs.

To overcome these limitations, we introduce \textit{BOOST}, a bootstrapping approach for few-shot reasoning program generation. The core insight of BOOST is to construct explicit meta-rules—generalizable guidelines that govern reasoning program construction. Rather than relying on manual examples, BOOST automatically derives these guidelines from limited labeled data through a data-driven critique-refine loop. This iterative process refines and generalizes the guidelines, leading to more effective and interpretable reasoning programs.

Specifically, BOOST is built upon a set of atomic functions that support flexible composition of sophisticated reasoning programs. By combining these atomic functions with automatically generated guidelines, BOOST systematically and efficiently produces high-quality few-shot demonstrations without human intervention.

Drawing inspiration from recent advances in agent optimization~\citep{agarwal2024promptwizard, NEURIPS2024_Trace}, BOOST uses program execution traces to deliver fine-grained, targeted feedback for guideline refinement, transcending mere verification correctness to avoid superficial enhancements. In each iteration, BOOST updates its guidelines based on this feedback, generates new program demonstrations, and evaluates them to select the most effective ones. This iterative process progressively improves demonstration quality without human supervision, enabling a seamless transition from zero-shot to few-shot learning for complex claim verification.

Our contributions are summarized as follows:
\begin{itemize}
    \item We propose BOOST, a bootstrapping approach for automated few-shot reasoning program generation. BOOST iteratively refines explicit, data-driven guidelines through a critique-refine loop, enabling a fully data-centric transition from zero-shot to few-shot learning without human intervention.

    \item We introduce a flexible set of atomic functions for reasoning program construction, enabling BOOST to generate more robust and adaptable reasoning programs.

    \item Experimental results on two benchmarks show that BOOST consistently outperforms existing approaches in both zero-shot and few-shot settings. The substantial performance gains achieved with bootstrapped demonstrations highlight the effectiveness of BOOST and represent a significant advancement in program-guided reasoning for fact-checking.

\end{itemize}

\section{Related Work}
\subsection{Claim Decomposition}
Claim decomposition is a key technique in explainable fact-checking and has seen widespread adoption~\citep{gunjal2024molecular, jiang2024core, song2024veriscore}. Many pipelines rely on LLMs for sub-claim generation via ICL~\citep{wanner-etal-2024-closer, wanner2024dndscore, kamoi-etal-2023-wice}. ProgramFC~\citep{pan-etal-2023-fact} adopts a program-guided approach for decomposition, while WICE~\citep{kamoi-etal-2023-wice} design prompts to extract atomic facts. Framing claim verification as question answering~\citep{chen-etal-2024-complex, ousidhoum-etal-2022-varifocal} has also shown effectiveness. However, \citet{hu2024decomposition} note that while decomposition simplifies reasoning, it also introduce noise, highlighting the need for adaptive strategies.

\subsection{Program-Guided Reasoning}  
Program-guided reasoning refers to techniques where a model utilizes explicit programs or structured procedures to guide its reasoning process. Instead of relying solely on free-form text reasoning, the model generates a symbolic plan (e.g., code, logic rules) that can be executed or evaluated~\citep{gao2023pal}. In numerical reasoning, Program-of-Thoughts (PoT)~\citep{chen2022program} prompts models to generate programs (e.g., in Python), allowing them to delegate arithmetic and logical operations to reliable computational tools. ProgramFC~\citep{pan-etal-2023-fact} was the first to apply this paradigm in fact-checking, transforming claims into executable reasoning programs, while FOLK~\citep{wang-shu-2023-explainable} proposed converting claims into First-Order Logic clauses. These methods build on chain-of-thought (CoT)~\citep{wei2022chain} prompting but introduce a layer of formal execution to reduce errors and enhance overall reasoning capabilities.

\section{Problem Definition}
Our problem definition aligns with the \textit{Open-book} setting introduced in ProgramFC~\citep{pan-etal-2023-fact}, which assumes access to a large textual corpus $\mathcal{K}$, such as Wikipedia. Given a claim $c$, a fact-checking system retrieves relevant \textit{evidence} from $\mathcal{K}$ and eventually predicts a binary veracity label $y$ (\textsc{True} or \textsc{False}). 

We focus on the \textit{Open-book} setting, as it closely reflects real-world fact-checking scenarios, where human fact-checkers must gather relevant evidence from extensive knowledge bases, without access to pre-compiled ground-truth evidence.

\section{Atomic Functions}
Designing atomic functions enhances flexibility in function composition and enables more diverse reasoning programs. ProgramFC introduces a set of sub-task functions; however, its design entangles multiple capabilities, such as enforcing retrieval to use question as the query within the QA function, limiting query formulation and evidence composition. Additionally, it includes redundant sub-task functions, such as explicitly evaluating logical expressions to produce boolean labels, even though the reasoning program inherently encodes logical reasoning steps. Since the underlying LLM can only access these predefined functions, its flexibility is constrained; it cannot freely configure retrieval queries or compose retrieved evidence to verify sub-claims.

To overcome these limitations, BOOST includes a new abstraction of \emph{atomic functions}, which explicitly decouple the core operations of reasoning program synthesis. Unlike prior design, this design provides three orthogonal and composable primitives—\texttt{RETRIEVE}, \texttt{QUESTION}, and \texttt{VERIFY}—enabling LLM to flexibly formulate queries, compose evidence, and express complex reasoning strategies with greater transparency and control.

\paragraph{RETRIEVE}  
Given a query $q$, this function returns the retrieved documents $e$ as a concatenated string.

\noindent\textbf{Exp:}
\texttt{retrieve(q) -> str}  

\paragraph{QUESTION}  
Given a question $q$, an evidence context $e$, and a desired output format $f$, this function returns an answer formatted according to $f$. Supported formats include ``sentence'', ``entity'', ``temporal'', and ``numeric'', each enforcing specific answer constraints. The function is implemented by prompting the LLM\footnote{Unless otherwise stated, the default LLM is GPT-4.1-mini~\citep{gpt41minireport}.} to generate an answer based on the provided question and evidence. Format-specific instructions are included in the prompt to ensure the answer strictly adheres to the specified format.

\noindent\textbf{Exp:}
\texttt{question(q, e, f) -> str}  

\paragraph{VERIFY}  
This function takes a (sub-)claim $c$ and an evidence context $e$ as input and returns a boolean veracity label (\textsc{True}/\textsc{False}). It is implemented by prompting the LLM to generate a verification decision. To improve verification accuracy, we employ chain-of-thought prompting.  

\noindent\textbf{Exp:}
\texttt{verify(claim, evidence) -> bool}

This design enhances modularity, improves flexibility in function composition. For more implementation details, please refer to Appendix.

\section{BOOST}

\subsection{Overview}
In low-resource scenarios where annotated data is limited and no ground-truth reasoning program are available, BOOST enables automated few-shot demonstration construction through a data-centric bootstrapping process. Starting without any initial guidelines, BOOST operates iteratively on mini-batches of data. In each iteration, BOOST generates claim-specific critiques, abstracts these into generalizable terms, and refines the existing guidelines accordingly. The refined guidelines then guide the generation of few-shot reasoning program demonstrations. Finally, among the generated candidates, BOOST selects and bootstraps the best-performing demonstration set for the next round.

\subsection{Guidelines Refinement}
Guidelines refinement aims to extract more data-centric logic, which is then reflected in bootstrapped demonstrations to improve their quality. Recent studies suggest that leveraging a more powerful LLM\footnote{For critique and refinement, we use GPT-4.1~\citep{gpt41report} by default.} as optimizer to critique and refine prompts can enhance performance. However, for the critique-refine process to be effective, the critique must accurately identify flaws and provide precise feedback, ensuring focused improvements rather than general changes.  

Human-annotated data typically consists of a claim $c$, a veracity label $y^{*}$, and ground-truth evidence $e^{*}$. However, relying solely on label-prediction correctness as a feedback signal is often insufficient for generating meaningful, directional critiques to guide program generation. Specifically, distinguishing whether an error stems from a missing bridging fact or a retrieval failure is highly challenging when analyzing only the program code.

\begin{figure*}[t]
\centering
\includegraphics[width=0.95\textwidth]{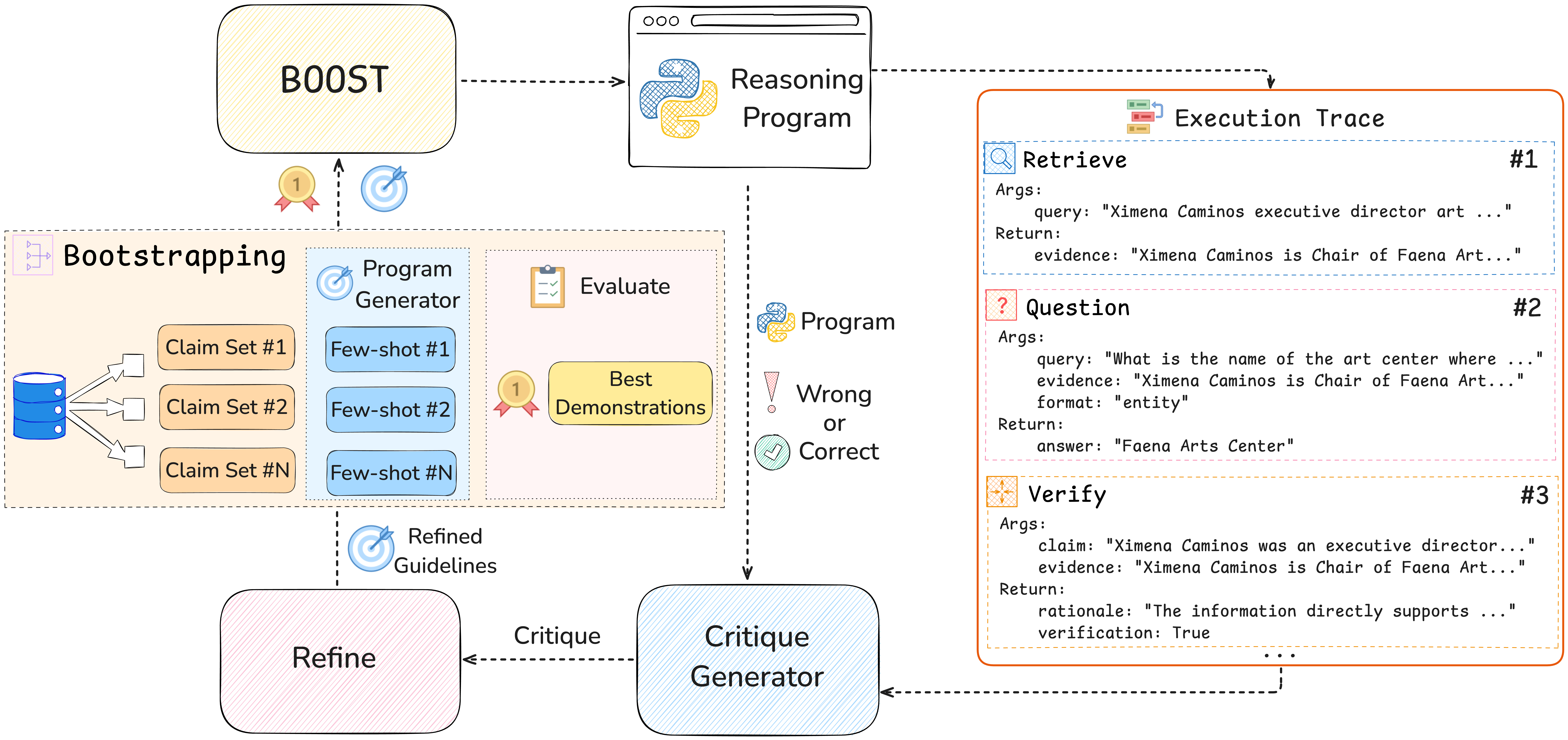}
\caption{Overview of the bootstrapping guideline-driven few-shot generation workflow. In each iteration, the reasoning program, label matching result, and execution trace are forwarded to generate a critique, which is then used to refine existing guidelines. The bootstrapping process samples $N$ claim sets and generates reasoning programs based on the updated guidelines. The generated demonstrations are evaluated using the F1 score.} 
\label{fig:bootstrap_workflow}
\end{figure*}

As shown in Figure~\ref{fig:bootstrap_workflow}, in order to generate meaningful and directional critiques, we introduce program execution trace as a key diagnostic signal.

\begin{algorithm}[t]
\caption{Bootstrapping Algorithm}
\begin{algorithmic}
\STATE \textbf{Input:} Total claim pool $C$, number of candidate claim sets $N$
\STATE \textbf{Output:} Refined guidelines $G^*$ and best few-shot demonstration set $S^*$

\STATE Initialize $G \gets None$, $S \gets \emptyset$, $C_{used} \gets \emptyset$, $Score^{*} \gets 0$
\STATE Set $G^* \gets G$, $S^* \gets S$

\FOR{each mini-batch $C_i$, $i = 2, \dots, n$}
    \STATE $G' \gets \text{GuidelinesRefinement}(G, C_i)$
    \STATE Sample $N$ claim sets: $\{C_{i,1}, C_{i,2}, ..., C_{i,N}\}$

    \FOR{each claim set $C_{i,j}$, $j = 0, \dots, N$}
        \STATE $S_{i,j} \gets \text{ProgramGeneration}(G', C_{i,j})$
        \STATE $Score_{i,j} \gets \text{Evaluate}(G', S_{i,j})$
    \ENDFOR

    \STATE Select best-performing demonstration set: 
    \STATE $S' \gets \arg\max_{j} \{ Score_{i,1}, ..., Score_{i,N} \}$

    \IF{$\max_j Score_{i,j} > Score^{*}$} 
        \STATE $Score^{*} \gets \max_j Score_{i,j}$
        \STATE $G^* \gets G'$, $S^* \gets S'$
    \ENDIF
\ENDFOR

\STATE \textbf{Return} $G^*, S^*$

\end{algorithmic}
\end{algorithm}

\paragraph{Program Execution Trace}: We track the execution of the reasoning program by recording each function call, its input arguments, and corresponding outputs in sequence: $T=[f_1(i_1) \rightarrow o_1, f_2(i_2) \rightarrow o_2, \dots]$. Since our \texttt{VERIFY} function is implemented with chain-of-thought prompting, we can capture the reasoning rationale for each verification step within the execution trace. Analyzing this trace helps precisely pinpoint where erroneous function calls or false verifications occur due to failed evidence retrieval, providing valuable signals for refining the information-gathering strategy.

Examples of the refined guidelines can be found in Appendix.

\subsection{Few-shot Generation with Bootstrapping}
Following guideline refinement, BOOST iteratively bootstraps claim sets and employs a program generator to produce reasoning program demonstrations that adhere to the updated guidelines. With the refined guidelines provided as context, the generator performs zero-shot reasoning to construct the corresponding reasoning programs. Beginning with an empty guideline $G$, BOOST repeatedly generates batch critiques to refine $G$ and simultaneously updates the few-shot program demonstrations $S$ to reflect the improved logic. This iterative bootstrapping process enables the automatic synthesis of high-quality demonstrations without human intervention.

At each iteration $i$, BOOST bootstraps $N$ candidate claim sets, each with $k$ claims, and generates the corresponding demonstration sets $S_{i,1}, \dots, S_{i,N}$. Bootstrapping multiple candidates enables broader exploration, reduces bias from individual claim selection, and increases the robustness of the resulting demonstrations. Each candidate set is then paired with the updated guidelines $G'$ and evaluated using an F1-score metric ($\mathrm{Score}$) on a held-out validation set. The highest-scoring set is selected as the few-shot demonstration set for further refinement.

This iterative selection over multiple mini-batches helps identify the most representative demonstration set~\citep{agarwal2024promptwizard}, enabling a data-centric and guideline-driven transition from zero-shot to few-shot learning while maintaining data efficiency and interpretability. An example of the generated demonstration can be found in the Appendix.

\begin{table*}[ht]
\centering
\begin{tabular}{l|cccccc|cccccc}
\toprule
& \multicolumn{6}{c}{\textbf{HOVER}} & \multicolumn{6}{c}{\textbf{FEVEROUS}} \\
\cmidrule(lr){2-7} \cmidrule(lr){8-13}
\textbf{Method} & \multicolumn{2}{c}{\textbf{2hop}} & \multicolumn{2}{c}{\textbf{3hop}} & \multicolumn{2}{c}{\textbf{4hop}} & \multicolumn{2}{c}{\textbf{Numerical}} & \multicolumn{2}{c}{\textbf{Multihop}} & \multicolumn{2}{c}{\textbf{Disambiguation}} \\
& F1 & Acc & F1 & Acc & F1 & Acc & F1 & Acc & F1 & Acc & F1 & Acc \\
\midrule
ProgramFC & 69.16$^{*}$ & 69.50$^{*}$ & 60.94$^{*}$ & 61.00$^{*}$ & 56.26$^{*}$ & 57.00$^{*}$ & 61.35$^{*}$ & 62.00$^{*}$ & 76.96$^{*}$ & 77.00$^{*}$ & 78.96$^{*}$ & 78.98$^{*}$ \\
QACheck & 70.49$^{*}$ & 70.50$^{*}$ & 58.90$^{*}$ & 59.00$^{*}$ & 56.41$^{*}$ & 56.50$^{*}$ & 66.70$^{*}$ & 67.50$^{*}$ & 70.00$^{*}$ & 70.00$^{*}$ & 74.88$^{*}$ & 75.00$^{*}$ \\
FOLK & 63.64$^{*}$ & 64.00$^{*}$ & 60.28$^{*}$ & 61.50$^{*}$ & 53.96$^{*}$ & 56.50$^{*}$ & 47.21$^{*}$ & 50.00$^{*}$ & 65.40$^{*}$ & 65.50$^{*}$ & 73.59$^{*}$ & 73.86$^{*}$ \\
BiDeV & 70.89$^{*}$ & 71.00$^{*}$ & 61.86$^{*}$ & 62.00$^{*}$ & 57.26$^{*}$ & 57.50$^{*}$ & 63.51$^{*}$ & 64.50$^{*}$ & 72.47$^{*}$ & 72.50$^{*}$ & 78.92$^{*}$ & 78.98$^{*}$ \\
VeGraph & 75.47$^{*}$ & 75.50$^{*}$ & \textbf{70.00} & \textbf{70.00} & 62.34$^{*}$ & 62.50$^{*}$ & 62.17$^{*}$ & 64.00$^{*}$ & 79.41$^{*}$ & 79.50$^{*}$ & 79.22$^{*}$ & 79.55$^{*}$ \\
\midrule
\textit{BOOST$_{\text{base}}$} & 75.99$^{*}$ & 76.00$^{*}$ & 65.35$^{*}$ & 65.50$^{*}$ & 61.02$^{*}$ & 62.50$^{*}$ & 70.70$^{*}$ & 71.50$^{*}$ & 78.16$^{*}$ & 78.50$^{*}$ & 78.00$^{*}$ & 78.41$^{*}$ \\
\textit{BOOST$_{\text{zs}}$} & 77.96$^{*}$ & 78.00$^{*}$ & 67.79 & 68.00 & 62.72$^{*}$ & 63.50$^{*}$ & 71.63$^{*}$ & 72.50$^{*}$ & 80.32$^{*}$ & 80.50$^{*}$ & 81.07 & 81.25 \\
\textit{BOOST$_{\text{fs}}$} & \textbf{79.97} & \textbf{80.00} & 69.33 & 69.50 & \textbf{65.50} & \textbf{66.00} & \textbf{75.11} & \textbf{75.50} & \textbf{85.47} & \textbf{85.50} & \textbf{82.26} & \textbf{82.39} \\
\bottomrule
\end{tabular}
\caption{Comparison of BOOST with baseline methods on HOVER and FEVEROUS benchmarks. F1 and accuracy (Acc) are reported for each partition.\protect\footnotemark[3]}
\label{tab:main-results}
\end{table*}
\footnotetext[3]{We mark significance ($p < 0.05$), based on 10 rounds of bootstrapped sampling~\citep{liu2025bidev}, with an asterisk (*). Scores that are significantly lower than those of \textit{BOOST$_{\text{fs}}$} are marked.}

\section{Experiments}
\subsection{Datasets \& Metrics}

We conduct experiments on two widely used complex claim verification benchmarks: HOVER and FEVEROUS. HOVER~\citep{jiang-etal-2020-hover} is designed for multi-hop claim verification, with supporting evidence scattered across multiple Wikipedia pages and includes 2-hop, 3-hop, and 4-hop partitions. The HOVER corpus is constructed from the October 2017 Wikipedia dump, using only the introductory sections of articles. FEVEROUS~\citep{Aly21Feverous} targets various claim verification types with both structured and unstructured data, and is based on the December 2020 Wikipedia dump with full article content. Consistent with prior work~\citep{Aly21Feverous, pham-etal-2025-verify}, we evaluate on three key FEVEROUS partitions: Multi-hop Reasoning, Numerical Reasoning, and Entity Disambiguation. Following established protocols~\citep{pham-etal-2025-verify, wang-shu-2023-explainable}, we sample 200 claims from each partition using stratified sampling to ensure balanced label distributions. Macro F1-score and accuracy are reported as the primary evaluation metrics. For the mini-batch update setup in each partition, each iteration operates on a batch of 10 examples, drawn from 100 sampled annotations.\protect\footnotemark[4]

\footnotetext[4]{We sample from the training data of our evaluated benchmarks to prevent data leakage.}

\subsection{Experimental Settings}
For all experiments, we use GPT-4.1-mini~\citep{gpt41minireport} as the underlying large language model. We adopt the \textit{open-book} setting~\citep{pan-etal-2023-fact}, where no ground-truth evidence is provided in advance. For all retrieval operation, we employ a two-stage system~\citep{pham-etal-2025-verify}: first, we use BM25-based sparse retrieval to obtain the top-50 candidate documents, then apply `bge-reranker-v2-m3' reranker~\citep{chen2024bge} to select the final top-5 documents. 
We evaluate BOOST in 3 settings:

\begin{itemize}
    \item \textit{BOOST$_{\text{base}}$}: Zero-shot without guidelines.
    \item \textit{BOOSTC$_{\text{zs}}$}: Zero-shot with guidelines refinement.
    \item \textit{BOOST$_{\text{fs}}$}: Few-shot with guidelines refinement and synthesized demonstrations.
\end{itemize}

\subsection{Baselines}
For baselines, we select the following approach related to explainable fact-checking pipelines:

\textbf{ProgramFC}~\citep{pan-etal-2023-fact}: Program-guided reasoning for claim verification, generating and executing reasoning programs in a few-shot manner. Serves as the primary program-guided reasoning baseline for comparison.

\textbf{QACheck}~\citep{pan-etal-2023-qacheck}: Verifies claims through iterative question answering, allowing the LLM to determine when sufficient information has been gathered. We evaluate the default Retriever–Reader setting, where the LLM answers questions using the corpus.

\textbf{FOLK}~\citep{wang-shu-2023-explainable}: Translates claims into First-Order Logic (FOL) clauses and applies FOL-guided reasoning over knowledge-grounded question–answer pairs. QA pairs are grounded via Google Search API.

\textbf{BiDeV}~\citep{liu2025bidev}: Proposes two prompt-based LLM agents—one for clarifying latent information, and another for removing redundant evidence—to address vagueness and redundancy in claims.

\textbf{VeGraph}~\citep{pham-etal-2025-verify}: Models claims as interactive graphs, iteratively resolving entity ambiguities and verifying sub-claims using LLM agents.

More experimental details can be found in Appendix.

\subsection{Main Results}
Table~\ref{tab:main-results} presents the fact-checking performance across different methods on the HOVER and FEVEROUS benchmarks. The results supports the following points:
\begin{itemize}
    \item \textit{BOOST$_{\text{fs}}$} achieves the best performance across both benchmarks and five partitions. On average, \textit{BOOST$_{\text{fs}}$} outperforms the strongest baseline by 4.09\% in F1 and 4.06\% in accuracy. The improvement is particularly notable for numerical reasoning, with an 8.41\% F1 gain over QACheck. Compared to the prior program-guided method ProgramFC, the average F1 improvement is 9\%. These results indicate the strong effectiveness of \textit{BOOST}.

    \item \textit{BOOST$_{\text{zs}}$} improves over \textit{BOOST$_{\text{base}}$}, and \textit{BOOST$_{\text{fs}}$} further surpasses \textit{BOOST$_{\text{zs}}$}. Compared to \textit{BOOST$_{\text{base}}$}, \textit{BOOST$_{\text{zs}}$} achieves a 1.84\% F1 gain, while \textit{BOOST$_{\text{fs}}$} delivers a 4.74\% improvement. These results highlight the effectiveness of guideline refinement and few-shot demonstration generation via bootstrapping.

    \item Atomic function design benefits program-guided reasoning. Compared to the previous program-guided reasoning method ProgramFC, \textit{BOOST$_{\text{base}}$}—even without guideline refinement and in a zero-shot setting—achieves a 4.27\% improvement in F1. This demonstrates the advantage of adopting a flexible set of atomic functions for reasoning program construction.
    
\end{itemize}

\begin{table}[t]
\centering
\begin{tabular}{lcccccc}
\toprule
& \multicolumn{2}{c}{\textbf{w/o critique}} & \multicolumn{2}{c}{\textbf{w critique}} \\
\cmidrule(lr){2-3} \cmidrule(lr){4-5}
\textbf{ } & F1 & Acc & F1 & Acc \\
\midrule
HOVER-2hops        & 73.47 & 73.50 & 77.96 & 78.00 \\
HOVER-3hops        & 67.20 & 67.50 & 67.79 & 68.00 \\
HOVER-4hops        & 60.14 & 61.50 & 62.72 & 63.50 \\
Fev.-Numerical     & 69.57 & 70.50 & 71.63 & 72.50 \\
Fev.-Multihop      & 80.35 & 80.50 & 80.32 & 80.50 \\
Fev.-Disambig      & 77.93 & 78.41 & 81.07 & 81.25 \\
\bottomrule
\end{tabular}
\caption{Performance comparison of guideline refinement with and without critique generation.}
\label{tab:ablation-critique}
\end{table}

\begin{table}[t]
\centering
\begin{tabular}{lccc}
\toprule
\textbf{ } & \textbf{Base} & \textbf{Random} & \textbf{Ours} \\
\midrule
HOVER-2hops        & 75.99 & 76.49 & 79.97 \\
HOVER-3hops        & 65.35 & 64.94 & 69.33\\
HOVER-4hops        & 61.02 & 58.81 & 65.50 \\
Fev.-Numerical     & 70.70 & 70.03 & 75.11 \\
Fev.-Multihop      & 78.16 & 79.84 & 85.47 \\
Fev.-Disambig      & 78.00 & 79.28 & 82.26 \\
\bottomrule
\end{tabular}
\caption{F1 performance comparison between randomly generated demonstrations and few-shot demonstrations synthesized by BOOST.}
\label{tab:ablation-random-fs}
\end{table}

\begin{table}[t]
\centering
\begin{tabular}{lcccc}
\toprule
\textbf{Partition} & \multicolumn{2}{c}{\textbf{Fewshot Refinement}} & \multicolumn{2}{c}{\textbf{Ours}} \\
\cmidrule(lr){2-3} \cmidrule(lr){4-5}
& F1 & Acc & F1 & Acc \\
\midrule
HOVER-2hops        & 76.94 & 77.00 & 79.97 & 80.00 \\
HOVER-3hops        & 65.83 & 66.00 & 69.33 & 69.50 \\
HOVER-4hops        & 58.94 & 60.50 & 65.50 & 66.00 \\
Fev.-Numerical     & 69.11 & 70.00 & 75.11 & 75.50 \\
Fev.-Multihop      & 77.43 & 77.50 & 85.47 & 85.50 \\
Fev.-Disambig      & 81.03 & 81.25 & 82.26 & 82.39 \\
\bottomrule
\end{tabular}
\caption{Performance comparison between direct few-shot refinement and few-shot demonstrations synthesized by BOOST.}
\label{tab:ablation-direct-fs}
\end{table}

\begin{figure}[t]
\centering
\includegraphics[width=1\linewidth]{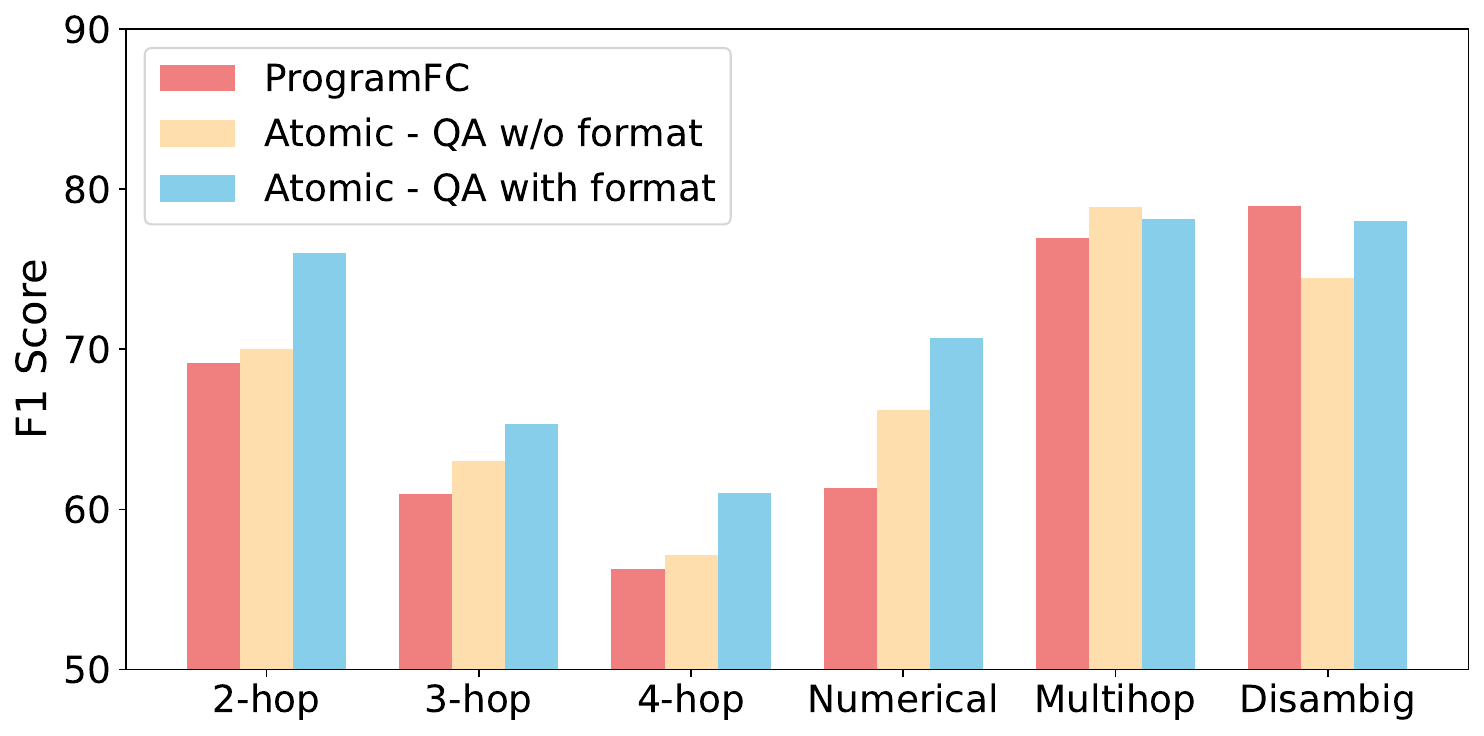}
\caption{F1 performance comparison of ProgramFC and BOOST using atomic functions, with and without format control in the QUESTION function.} 
\label{fig:atomic_func}
\end{figure}

\begin{figure}[!h]
\centering
\includegraphics[width=1\linewidth]{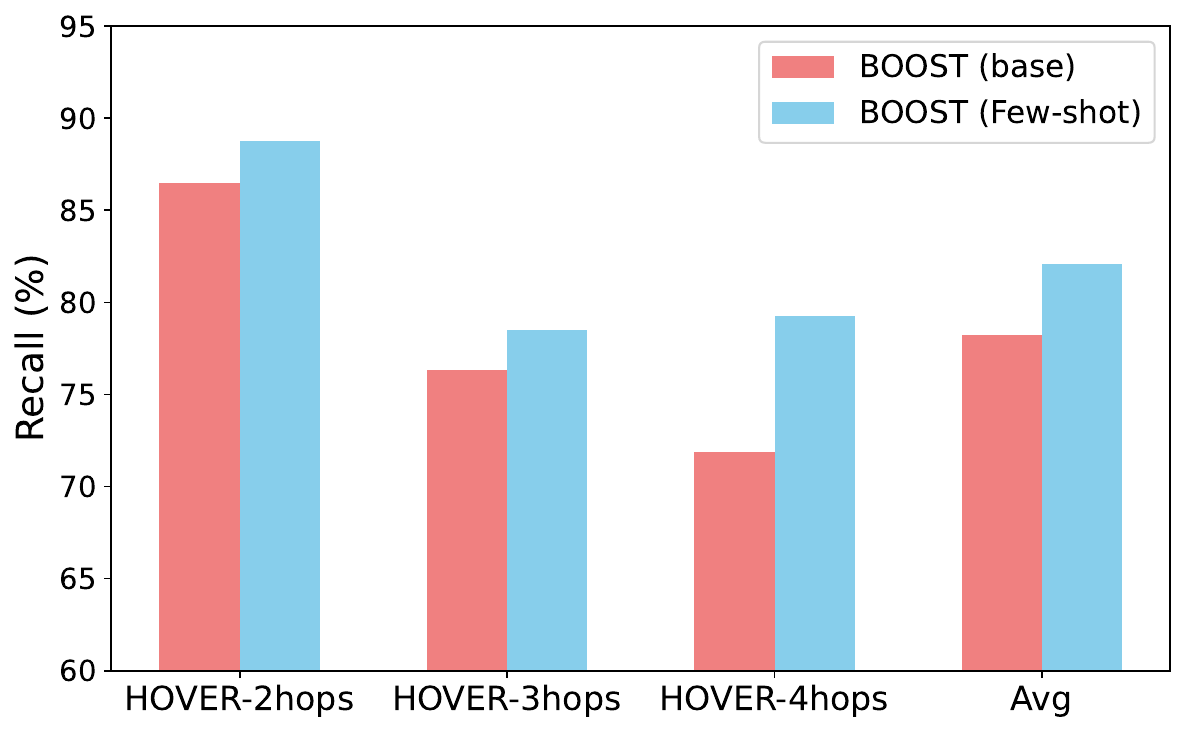}
\caption{Evidence recall comparison of \textit{BOOST$_{\text{base}}$} and the final \textit{BOOST$_{\text{fs}}$}.} 
\label{fig:evidence_recall}
\end{figure}

\subsection{Ablations and Analysis}

We conduct ablation and analysis experiments to examine the following aspects: (1) the impact of critique generation for guideline refinement, (2) comparison with demonstrations generated directly by LLM, (3) comparison with direct refinement of few-shot examples, (4) the effect of atomic function design, and (5) evidence recall rate.

\subsubsection{Impact of Critique Generation}
We compare our approach using critique generation for guideline refinement with using only sample correctness as feedback. As shown in Table~\ref{tab:ablation-critique}, incorporating critique generation generally improves verification performance. On the 2-hop and disambiguation partitions, the use of a critique generator yields F1 improvements of 4.49\% and 3.1\%, respectively. These results suggest that our critique generation, which leverages program execution traces to provide more fine-grained and actionable feedback, enables more effective refinement of guidelines and overall performance gains.

\subsubsection{Comparison with Generated Few-Shot}
We compare our bootstrapping-based, guideline-driven approach for few-shot synthesis with randomly generated demonstrations. Specifically, we randomly sample three claims from each benchmark partition and prompt GPT-4.1 to generate the corresponding reasoning program demonstrations. As shown in Table~\ref{tab:ablation-random-fs}, our method consistently outperforms the random baseline by an average of 4.61\% in F1. Notably, randomly generated demonstrations yield marginal improvements and can even lead to performance degradation on 4-hop data. These results underscore the effectiveness of BOOST.

\begin{figure*}[ht]
\centering
\includegraphics[width=0.95\linewidth]{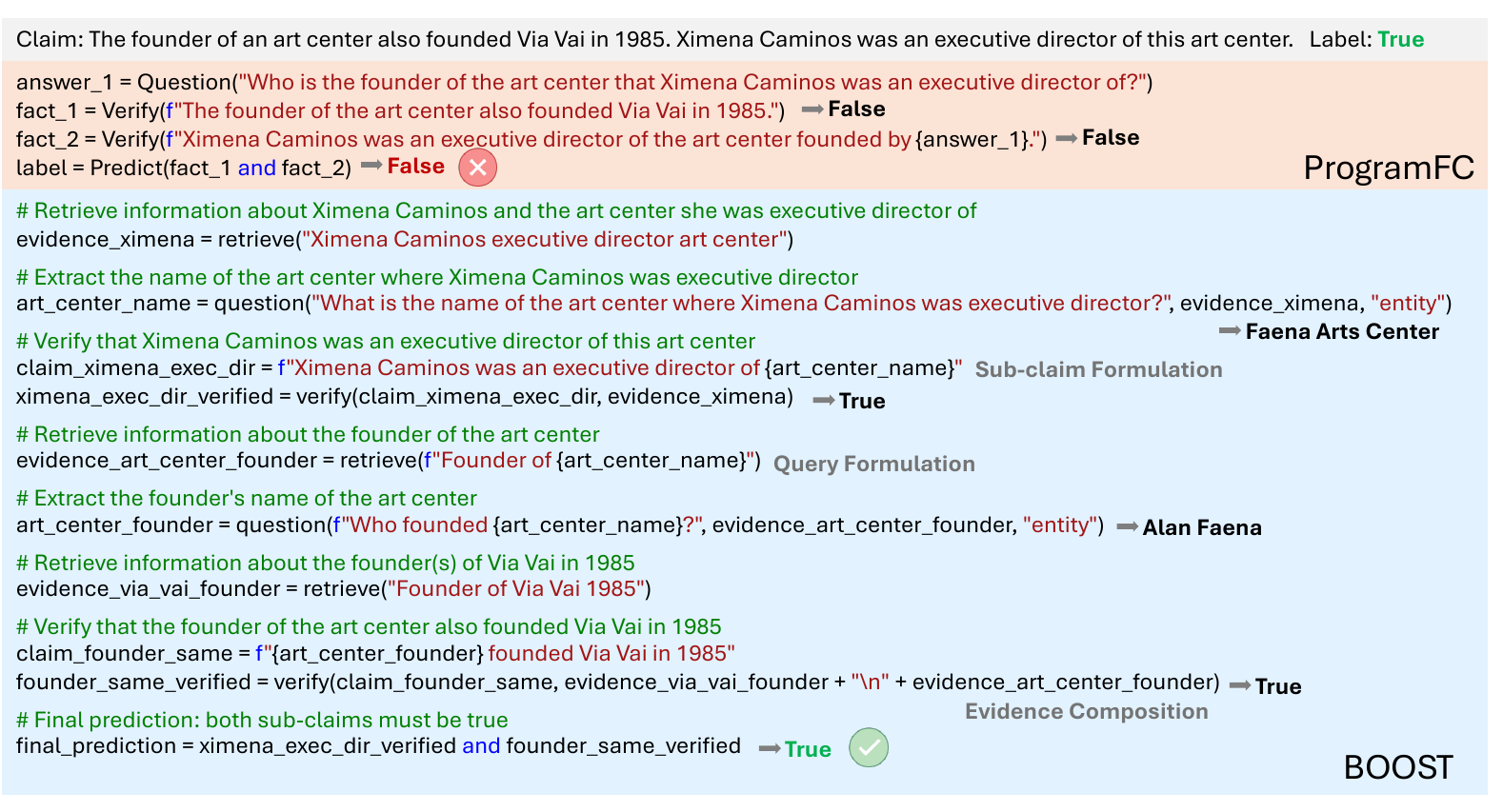}
\caption{Case study comparing the reasoning programs generated by ProgramFC and BOOST.} 
\label{fig:case_study}
\end{figure*}

\subsubsection{Comparison with Direct Few-Shot Refinement}
BOOST synthesizes effective few-shot demonstrations through iterative guideline refinement. We compare this approach to direct few-shot refinement, where few-shot examples are initialized using randomly generated reasoning programs. Instead of generating critiques to refine guidelines at each step, the baseline method directly refines each demonstration in the provided few-shot context. As shown in Table~\ref{tab:ablation-direct-fs}, our method achieves substantial improvements over direct few-shot refinement, with an average F1 increase of 4.89\%. This result highlights the importance of explicit guideline refinement in producing high-quality demonstrations.

\subsubsection{Effect of Atomic Function Design}
The previous method, ProgramFC, entangles evidence retrieval within its \texttt{VERIFY} and \texttt{QUESTION} functions, enforcing the use of the question or claim as the query. This restricts query formulation and evidence composition. In contrast, BOOST is designed to leverage the LLM’s coding capabilities by encouraging more flexible combinations of atomic functions. Specifically, BOOST uses \texttt{RETRIEVE}, \texttt{QUESTION}, and \texttt{VERIFY} as distinct atomic functions. To further improve control, the \texttt{QUESTION} function in BOOST supports an explicit format attribute, addressing the uncontrolled output format issue of freeform question-answering. We compare the F1 performance of these different function designs in Figure~\ref{fig:atomic_func}. The results show that our atomic function design consistently yields the best performance.

\subsubsection{Evidence Recall}
In addition to verification performance, we evaluate the evidence-gathering capability of our approach by reporting the evidence recall of \textit{BOOST$_{\text{base}}$} and \textit{BOOST$_{\text{fs}}$}, as shown in Figure~\ref{fig:evidence_recall}. Recall measures the proportion of ground-truth evidence retrieved among all retrieved documents, reflecting the system’s effectiveness in evidence retrieval. The results confirm that our few-shot generation approach clearly enhances retrieval performance.

\section{Case Study}
To provide an intuitive illustration of BOOST’s advanced reasoning capabilities, we present a comparative case study against the previous program-guided reasoning approach, ProgramFC. As shown in Figure~\ref{fig:case_study}, ProgramFC produces an incorrect prediction due to its entangled function design. For example, in its first \texttt{VERIFY} call, it uses a multi-hop sub-claim as the query for both retrieval and verification, resulting in incomplete evidence retrieval and incorrect verification.

In contrast, BOOST decomposes the claim into finer-grained, interpretable sub-steps using distinct atomic functions. In this case, BOOST first isolates and extracts key entities—such as the art center and its founder—before verifying each sub-claim with targeted evidence. By separating retrieval and verification, BOOST ensures that each query is well-formed, resulting in more complete evidence gathering and a correct final prediction. Furthermore, BOOST demonstrates the ability to compose evidence from multiple sources using connectors such as `\texttt{\textbackslash n}', allowing it to integrate multiple pieces of retrieved information into a single, coherent verification step.

\section{Conclusion}
We introduce BOOST, a bootstrapping-based approach for few-shot reasoning program generation. BOOST explicitly incorporates data-centric guideline refinement, iteratively improving bootstrapped demonstrations in a guideline-driven manner. This enables a seamless transition from zero-shot to few-shot learning without human intervention, enhancing both interpretability and fact-checking effectiveness. Experimental results on two benchmarks demonstrate that BOOST outperforms prior few-shot baselines, showcasing its superior performance and the value of automated, interpretable program synthesis for complex reasoning tasks.

\bibliography{custom}

\end{document}